\documentclass{article}

\usepackage{subfig}
\usepackage{wrapfig}
\usepackage{microtype}
\usepackage{graphicx}
\usepackage{booktabs} 
\usepackage{multirow} 
\usepackage{soul}
\usepackage{xcolor,colortbl} 
\usepackage{changepage,threeparttable}
\usepackage[preprint]{neurips_data_2024}
\usepackage{amsmath}
\usepackage{amssymb}
\usepackage{mathtools}
\usepackage{amsthm}

\usepackage[utf8]{inputenc} 
\usepackage[T1]{fontenc}    
\usepackage{hyperref}       
\usepackage{url}            
\usepackage{booktabs}       
\usepackage{amsfonts}       
\usepackage{nicefrac}       
\usepackage{microtype}      
\usepackage{xcolor}         

\title{Inference-Time Computations for LLM Reasoning and Planning: A Benchmark and Insights}

\author{%
  Shubham Parashar\thanks{Co-first authors;  Project page at \href{https://github.com/divelab/sys2bench}{https://github.com/divelab/sys2bench}.}$^{*}$ \quad
  Blake Olson$^{*}$ \quad
  Sambhav Khurana$^{*}$ \quad
  Eric Li$^{*}$ \quad
  Hongyi Ling \\
  \textbf{James Caverlee} \quad
  \textbf{Shuiwang Ji} \\[5pt]
  Department of Computer Science \& Engineering\\
  Texas A\&M University \\[5pt]
}

\begin{document}

\maketitle

\begin{abstract}
We examine the reasoning and planning capabilities of large language models (LLMs) in solving complex tasks. Recent advances in inference-time techniques demonstrate the potential to enhance LLM reasoning without additional training by exploring intermediate steps during inference. Notably, OpenAI's o1 model shows promising performance through its novel use of multi-step reasoning and verification. 
Here, we explore how scaling inference-time techniques can improve reasoning and planning, focusing on understanding the tradeoff between computational cost and performance.
To this end, we construct a comprehensive benchmark, known as \emph{Sys2Bench}, and perform extensive experiments evaluating existing inference-time techniques on eleven diverse tasks across five categories, including arithmetic reasoning, logical reasoning, common sense reasoning, algorithmic reasoning, and planning. 
Our findings indicate that simply scaling inference-time computation has limitations, as no single inference-time technique consistently performs well across all reasoning and planning tasks. 
\end{abstract}

\section{Introduction}
\label{sec:intro}

Large language models (LLMs)~\citep{brown2020language} have demonstrated exceptional performance across a range of natural language processing (NLP) tasks, including question answering, machine translation, sentiment analysis, and text summarization~\citep{devlin2019bert,vaswani2017attention}. Beyond NLP, LLMs have also been adapted for multimodal tasks involving vision~\citep{parashar2024neglected, lin2025evaluating} and audio~\citep{wu2024towards}. Building on their success in these diverse domains, researchers are increasingly using LLMs as AI agents~\citep{deng2024mind2web, wang2024survey} for complex tasks, such as robotics~\citep{liu2023llmp} and scientific discovery~\citep{wang2024survey}. These tasks require the reasoning and planning capabilities of LLMs, extending beyond simpler text comprehension.

Reasoning and planning in LLMs refer to their ability to solve complex problems by understanding, processing, and generating solutions across various domains~\citep{hao2024llm}. These capabilities can be analyzed from multiple perspectives; we propose a classification that organizes reasoning and planning tasks into five categories, namely arithmetic, logical, commonsense, algorithmic, and plan generation challenges. 
Recent advances in inference-time techniques demonstrate the potential to enhance LLM reasoning and planning without additional training. These techniques focus on decomposing complex problems into simpler intermediate steps during inference. For instance, Chain-of-Thought~\citep{wei2022chain} encourages step-by-step reasoning, while Tree-of-Thought~\citep{yao2024tree} chooses optimal reasoning paths using tree search. Notably, OpenAI's O1 model, a large reasoning model (LRM)~\citep{canLrmPlan}, achieves state-of-the-art performance on various reasoning tasks, demonstrating the effectiveness of inference-time techniques. 
This success has inspired the research community to focus more on scaling inference-time techniques in the hope of similar performance improvements.

Although inference time techniques have improved LLM reasoning and planning, evaluation of these methods has been limited to specific tasks, models, and datasets. Moreover, these methods have additional computational costs, presenting a trade-off between computational overhead and performance gains. To overcome this limitation we introduce Sys2Bench, a comprehensive benchmark covering multiple tasks and models. 
Specifically, we perform experiments on eleven datasets and seven different LLMs, testing four widely used inference-time techniques. Based on our findings, we argue that simply scaling inference-time computation has limitations. Instead, we need to explore diverse approaches to enhance the holistic reasoning capabilities of LLMs, as no single inference-time technique consistently outperforms others across all tasks. 

\section{Related Work}
\label{sec:related}
\textbf{LLM Reasoning} is the ability of LLMs to logically process information and draw coherent conclusions, enabling them to solve complex problems~\citep{PrOntoQA}. The success of LLMs in Natural Language Generation~\citep{radford2018improving} and Natural Language Understanding~\citep{vaswani2017attention, devlin2019bert} has sparked interest in exploring reasoning capabilities. A range of datasets have been introduced to evaluate reasoning, covering tasks in arithmetic~\citep{aqua,cobbe2021gsm8k}, logic~\citep{arc_agi,wang2022lsat}, common sense~\citep{yang2018hotpotqa, geva2021did}, and algorithmic reasoning~\citep{yao2024tree}. We introduce these tasks in more detail in 
Section~\ref{sec:eval}, and report results across these tasks in Section~\ref{sec:experiments}.

\textbf{LLM Planning} involves constructing a sequence of actions to achieve defined goals~\citep{valmeekam2023planning, zheng2024natural}. LLMs have been employed as planners or high-level controllers for robotic tasks~\cite{liu2023llmp, huang2022language} and as agents for web navigation~\citep{deng2024mind2web}, scientific discovery~\citep{wang2024survey}, and autonomous vehicles~\citep{yang2023llm4drive}. Despite their broad adoption, studies reveal that LLMs often struggle to generate valid plans for complex  tasks~\citep{kambhampatiposition, xie2024travelplanner}. We provide details on evaluated planning problems in Section~\ref{sec:eval}, with results and analyses in Section~\ref{sec:experiments}.

\textbf{Inference Time Techniques} 
for LLMs are methods applied during output generation to improve performance, and alignment with downstream tasks~\citep{welleck2024from}. These techniques aid reasoning and planning by breaking complex tasks into smaller, manageable steps for systematic problem-solving. For instance, Chain-of-Thought prompting (CoT)~\citep{wei2022chain} and its variants~\citep{zhou2023leasttomost, kojima2022large} decompose problems into sequential steps, while self-consistency~\citep{wang2023selfconsistency} refines CoT by aggregating multiple responses through voting. Tree of Thought~\citep{yao2024tree}, Graph of Thought~\citep{besta2024graph}, and Monte Carlo Tree Search~\citep{rap,lats} enhance problem-solving by systematically exploring reasoning paths. Details on inference-time methods are in Section~\ref{sec:inference_time}, with results in Section~\ref{sec:experiments}.

\section{Sys2Bench Problems and Datasets}
\label{sec:eval}
\begin{table*}[]
\centering
\small
\caption{\small Summary of the 11 datasets included in Sys2Bench.}
\label{tab:dataset-desc}
\setlength{\tabcolsep}{2.0pt}
\scalebox{0.613}{
\begin{tabular}{@{}lllllll@{}}
\toprule
& \multicolumn{2}{c}{Algorithmic Reasoning} & \multicolumn{4}{c}{Planning} \\ \cmidrule(lr){2-3} \cmidrule(lr){4-7}  
 & Game of 24 & Binpacking & Blocksworld & Trip Plan & Calendar Plan & Rubik's Cube \\ \midrule 
Task & \begin{tabular}[c]{@{}l@{}}Propose an arithmetic \\ expression to reach 24.\end{tabular} & \begin{tabular}[c]{@{}l@{}}Pack items into the \\ fewest bins.\end{tabular} & \begin{tabular}[c]{@{}l@{}}Plan actions to transform\\ blocks from initial to goal state.\end{tabular} & \begin{tabular}[c]{@{}l@{}}Plan a trip across cities \\ for a set number of days.\end{tabular} & \begin{tabular}[c]{@{}l@{}}Schedule a meeting considering\\ time constraints of people.\end{tabular} & \begin{tabular}[c]{@{}l@{}}Unscramble a scrambled\\ 2×2 Rubik's Cube.\end{tabular} \\ \midrule
Input & A list of 4 numbers. & \begin{tabular}[c]{@{}l@{}}List of item weights\\ and bin capacity.\end{tabular} & \begin{tabular}[c]{@{}l@{}}Initial state of blocks \\ and goal state.\end{tabular} & \begin{tabular}[c]{@{}l@{}}Cities, days per city, total\\ days, and possible flights.\end{tabular} & \begin{tabular}[c]{@{}l@{}}Calendars with meetings and \\ time constraints.\end{tabular} & \begin{tabular}[c]{@{}l@{}}A scrambled 2×2 Rubik's\\ Cube.\end{tabular} \\ \midrule
Output & \begin{tabular}[c]{@{}l@{}}An arithmetic \\ expression.\end{tabular} & \begin{tabular}[c]{@{}l@{}}Final list with items \\ arranged in bins.\end{tabular} & \begin{tabular}[c]{@{}l@{}}A sequence of actions \\ as the plan.\end{tabular} & A trip itinerary. & \begin{tabular}[c]{@{}l@{}}A meeting time fitting all \\ schedules.\end{tabular} & \begin{tabular}[c]{@{}l@{}}A sequence of rotations\\ that unscramble the cube.\end{tabular} \\ \midrule
& \multicolumn{2}{c}{Arithmetic Reasoning} & \multicolumn{1}{c}{Logical Reasoning} & \multicolumn{2}{c}{Common Sense Reasoning} \\ \cmidrule(lr){2-3} \cmidrule(lr){4-4} \cmidrule(lr){5-6}
& GSM8K & AQuA & ProntoQA & StrategyQA & HotPotQA \\ \cmidrule(lr){1-6}
Task & \begin{tabular}[c]{@{}l@{}}Solve high school  \\ arithmetic problems.\end{tabular} & \begin{tabular}[c]{@{}l@{}}Solve algebraic \\ problems.\end{tabular} & \begin{tabular}[c]{@{}l@{}}Draw a logical conclusion\\ from a set of predicates.\end{tabular} & \begin{tabular}[c]{@{}l@{}}Answer general knowledge\\ questions.\end{tabular} & \begin{tabular}[c]{@{}l@{}}Answer general knowledge \\ questions using provided facts.\end{tabular} \\ \cmidrule(lr){1-6}
Input & \begin{tabular}[c]{@{}l@{}}Arithmetic problem\\ description.\end{tabular} & \begin{tabular}[c]{@{}l@{}}Algebraic problem\\ description.\end{tabular} & \begin{tabular}[c]{@{}l@{}}A clause to verify as true or \\ false using logical predicates.\end{tabular} & A yes/no question. & \begin{tabular}[c]{@{}l@{}}General knowledge question\\ with supporting facts.\end{tabular} \\ \cmidrule(lr){1-6}
Output & A numerical value. & A multiple-choice option. & True or False, with reasoning. & Yes or No. & Short answer of 1 or 2 words. \\ \bottomrule
\end{tabular}
}
\end{table*}

In this section, we introduce \textbf{\emph{Sys2Bench}}, a benchmark designed to systematically evaluate the reasoning and planning capabilities of Large Language Models (LLMs) across diverse tasks. The name Sys2Bench reflects its focus on evaluating Systematic Reasoning and Planning, providing a structured framework for assessing inference-time techniques.

A key motivation for this benchmark is to \emph{demonstrate the limitations of simply scaling inference-time computation}, showing that it does not consistently lead to better reasoning or problem-solving abilities. While inference-time techniques have gained traction in improving LLM performance, no single approach consistently outperforms others across all tasks. Thus, we argue that a more holistic exploration of reasoning strategies is essential. Sys2Bench facilitates this by benchmarking LLMs on eleven datasets, categorized into five primary reasoning types: Arithmetic Reasoning, Logical Reasoning, Common Sense Reasoning, Algorithmic Reasoning, and Planning (summarized in Table ~\ref{tab:dataset-desc}). 

\subsection{Arithmetic Reasoning}
The ability of Large Language Models (LLMs) to solve multi-step arithmetic problems remains an active area of research~\cite{snell2024scaling, kumar2024training, hendrycks2021measuringMATH}.  Additionally, OpenAI's o1 models ~\citep{openai2024o1systemcard} have prompted the research community to explore inference-time techniques to improve the arithmetic reasoning of LLMs~\citep{zhao2024marco}. We evaluate the arithmetic reasoning of LLMs, on  \textbf{GSM8K}~\citep{cobbe2021gsm8k} and \textbf{AQuA}~\citep{aqua} benchmark. 

\textbf{GSM8K} is a popular dataset of high-quality, linguistically diverse elementary school math word problems, designed to evaluate multi-step arithmetic reasoning. The problems typically require 2 to 8 steps of arithmetic operations, testing the ability of LLMs to perform logical deduction and basic calculations.

\textbf{AQuA} is a dataset of around 100,000 algebraic word problems with multiple-choice answers and detailed rationales. It is designed to evaluate the arithmetic reasoning and problem-solving capabilities of models, making it a challenging benchmark for LLMs.

\subsection{Logical Reasoning}
Logical reasoning involves deriving conclusions based on a structured sequence of rules, or premises. The evaluation of the ability to reason logically by LLM helps assess their ability to solve structured and complex decision-making problems~\citep{arc_agi}. We use \textbf{ProntoQA}~\citep{PrOntoQA} to evaluate the logical reasoning ability of LLMs.

\textbf{ProntoQA} is a dataset developed to evaluate an LLM's ability to reason and generate explicit reasoning chains for first-order logic-based queries~\citep{barwise1977introduction}. It challenges models to not only produce correct answers but also provide detailed, step-by-step reasoning paths that justify their conclusions.

\subsection{Common Sense Reasoning} Common Sense Reasoning is the process of drawing conclusions from implicit everyday knowledge. Evaluating this skill ensures that LLMs provide accurate and contextually appropriate responses. We evaluate this type of reasoning using the \textbf{StrategyQA}~\cite{geva2021did} and \textbf{HotPotQA}~\cite{yang2018hotpotqa} datasets.


\textbf{StrategyQA} is a benchmark designed to assess a model's ability to perform implicit multi-step reasoning using general knowledge or common sense facts. It consists of yes/no questions where the goal is to arrive at the correct answer by generating and verifying intermediate reasoning steps.

\textbf{HotPotQA} is a large-scale dataset designed to evaluate how effectively models combine information from multiple documents to answer general knowledge questions. It features diverse question types and tests the use of sentence-level evidence for accurate and explainable multi-hop reasoning.

\subsection{Algorithmic Reasoning}
We focus on applying LLMs to solve complex NP-hard and NP-complete tasks, requiring them to evaluate constraints and propose optimized algorithms that achieve practical and effective solutions. Such problems assess the application of LLMs to combinatorial optimization and resource allocation tasks~\citep{liu2024evolution,romera2024mathematical}. We use \textbf{Game of 24}~\citep{yao2024tree}, and a novel dataset, \textbf{Bin Packing}.

\textbf{Game of 24} is a dataset where the goal is to form an arithmetic expression evaluating to 24 using '+', '-', '*', or '/' with a list of four numbers. As an NP-complete problem with multiple solutions, it challenges an LLM to efficiently generate expressions by focusing only on operations that can lead to the target value.

\textbf{Bin Packing} is a new task introduced by us, inspired by the combinatorial optimization problems studied by~\citet{liu2024evolution, romera2024mathematical}. In this task, the goal is to find the least number of bins needed to pack a list of items. Specifically, a list of $N$ items of weight $[ W_1, W_2, ... W_n ]$ is given, which must be divided into bins $B_1, B_2, B_3...B_m$. The sum of weights in each bin must not exceed the bin capacity 
$C$, and the objective is to minimize the total number of bins $m$. Formally, the task can be written as: 
\vspace{-.5cm}
{
\setlength{\abovedisplayskip}{0pt}

\begin{equation}
\scalebox{0.8}{$
\begin{aligned}
\min \ \  m \quad \text{subject to} \quad 
\begin{cases} 
& \bigcup_{j=1}^m B_j = \{1,\dots,n\}, \quad B_j \cap B_{j'} = \varnothing \quad (\forall j \neq j'), \\\\
& \sum_{i \in B_j} W_i \le C \quad (\forall j).
\end{cases}
\end{aligned}
$}
\end{equation}
}


\subsection{Planning} 
A planning problem is defined by $(S_0,A,G)$, where $S_0$ stands for an initial state, $A$ is the set of actions needed to achieve the goal $G$. Planning problems require LLMs to demonstrate multistep reasoning, and sound decision making to arrive at correct solutions. These problems have broad applications in robotics and agent-based systems. Our evaluation focuses on four planning problems: BlocksWorld~\citep{valmeekam2023planning}, Rubik's Cube~\citep{everything-of-thoughts}, TripPlan, and CalendarPlan~\citep{zheng2024natural}.

\textbf{BlocksWorld} is a popular dataset to evaluate the planning capabilities of LLMs. Each task involves transitioning from an initial block configuration to a target configuration, which requires LLMs to generate a sequence of actions to achieve the goal.

\textbf{Rubik's Cube} requires an LLM to solve a scrambled $2 \times2$ cube by restoring each face to a uniform color. Starting from a scrambled cube, the LLM must generate a valid plan of cube rotations to achieve the goal. 

\textbf{Trip Plan} challenges an LLM to plan a travel itinerary that satisfies constraints on cities, dates, and flight connectivity, ensuring that all cities are visited as specified. 

\textbf{Calendar Plan} is a dataset designed to schedule a meeting by aligning the availability of a group of people. The goal is to find a feasible time slot that accommodates all the constraints of the participants. 

\section{Sys2Bench Baseline Methods}
\label{sec:inference_time}

\begin{figure*}[t]
    \centering
    \includegraphics[width=\textwidth]{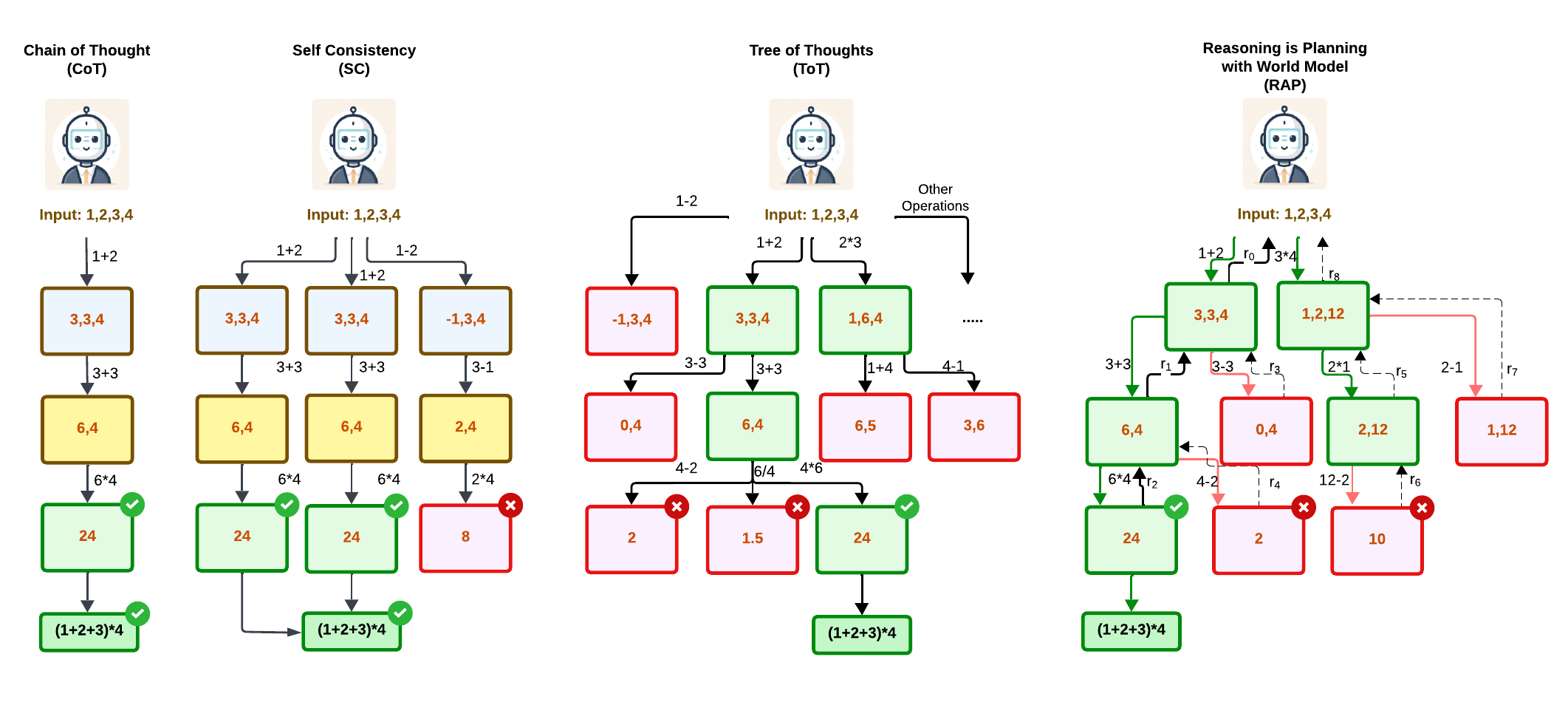} 
    \caption{ \small Overview of Inference-Time Techniques evaluated on the Game of 24 dataset. We evaluate four inference-time reasoning techniques. Chain of Thought (CoT)~\citep{wei2022chain} solves problems through a linear sequence of reasoning steps. Self-Consistency (SC)~\citep{wang2023selfconsistency} extends CoT by selecting answers through majority voting over multiple reasoning chains. Tree of Thoughts (ToT)~\citep{yao2024tree} uses tree search to explore and expand reasoning paths. Reasoning as Planning (RAP)~\cite{rap} combines Monte Carlo Tree Search (MCTS) with the LLM as a world model to reward reasoning steps and guide tree growth toward the answer.}
    \label{fig:test_time}
\end{figure*}

In Sys2Bench we evaluate popular inference-time techniques commonly used to enhance System 2 abilities of LLMs. While these techniques have typically been applied to specific tasks, we analyze their performance comprehensively in Sys2Bench. 
Sys2Bench allows us to uncover patterns and limitations that may not be previously evident. We summarize these methods in Fig.~\ref{fig:test_time}.

\textbf{Chain of Thought} (CoT) enables LLMs to solve complex problems by breaking them into intermediate reasoning steps, improving their logical coherence and accuracy~\cite{wei2022chain}. CoT enhances structured problem-solving of LLMs by providing in-context examples of step-by-step reasoning during inference.

\textbf{Self Consistency} (SC) extends CoT by generating multiple reasoning paths for a problem and selecting the most consistent answer through majority voting~\citep{wang2023selfconsistency}.

\textbf{Tree of Thoughts} (ToT) uses structured tree search to enhance reasoning in LLMs by systematically exploring multiple paths, with the LLM evaluating its own intermediate generations to decide which paths to expand~\citep{yao2024tree}. Evaluation can be performed by rating LLM generation on a scale of 1-10 or using logits for scoring.

ToT has three search strategies: depth-first search (DFS), breadth-first search (BFS), and beam search. In our experiments, we use beam search because it performs the best amongst all variants.

\textbf{Reasoning as Planning with World Models} (RAP) reformulates reasoning as a planning problem, where the LLM acts as both the reasoning agent and the world model~\citep{rap}. The reasoning agent generates potential reasoning paths, while the world model simulates and evaluates these paths. Specifically, RAP uses Monte Carlo Tree Search (MCTS)~\citep{coulom2006efficient} to explore and refine reasoning paths.

Unlike ToT, which does exhaustive tree search, RAP dynamically prioritizes high-potential paths using MCTS, resulting in improved performance. RAP requires logits for MCTS, which is why it is exclusively implemented on LLaMA. Since RAP requires extensive prompt engineering to frame all tasks as planning problems, we evaluate it on a subset of tasks, including GSM8K, AQuA, ProntoQA, StrategyQA, Game of 24, Binpacking, Blocksworld, and Rubik's Cube.


\section{Experiments}
\label{sec:experiments}
In this section, we present the experiments conducted on various tasks in the Sys2Bench benchmark. We begin by outlining the experimental setup, detailing the models and implementation specifics of the inference-time methods. Next, the results for the different inference-time methods are shown in Table~\ref{tab:results-big} and Table~\ref{tab:o1-results}. 
\subsection{Setup}
In this subsection, we provide details about the experimental setup used to evaluate the performance of various inference-time techniques in Sys2Bench. We describe the models, the implementation specifics of the inference-time methods, and the metrics used for evaluation.

\textbf{Models} evaluated in Sys2Bench, consist of three LLaMa 3.1 models, two GPT-based models, and two large reasoning models (LRMs). The LLaMa 3.1 variants are 8B, 70B, and 405B, while the GPT-based models include GPT-4o and GPT-4o-mini. Additionally, the O1 and O1-mini models are tested as part of our LRM evaluation. By default, we use a temperature of 0.8 across all models for generation.


\textbf{Chain of Thought (CoT)} involves including in-context learning examples in the prompt. In our benchmark, we limit this to five examples per prompt. These examples are selected from the in-context examples provided by the dataset or the training set. If neither is available, we use a subset of test examples and evaluate the remaining test instances.  

\textbf{Self Consistency (SC)} follows the same settings as CoT. We generate five CoT responses from the LLM and determine the final output through majority voting. 

\textbf{Tree of Thought (ToT)} implementation in Sys2Bench uses beam search. The beam size is 5 for most tasks except planning tasks, where the number of possible actions at each state is larger, thus, the beam size is increased to 10. By default, beam ranking is performed by asking the LLM to rate outputs on a scale of 1 to 10, except for LLaMa 3.1 8B, where logits are used instead. Finally, the search depth is task-dependent, ranging from 4 for Game of 24 to 20 for Trip Plan. 

\textbf{Reasoning as Planning (RAP)} uses Monte Carlo Tree Search (MCTS) with up to 10 rollouts during inference. Due to its reliance on a reward model that requires logits, RAP is implemented exclusively on LLaMa 3.1 8B. Similar to ToT, the search depth in RAP varies depending on the task. 

\textbf{Input Output Prompting (IO)} is utilized with LRMs, as they generate their own reasoning steps and do not require in-context learning examples. Instead, we provide the necessary format and instruct the models to respond in the same format.

\textbf{Metric} used across all tasks is accuracy. Note that the context of accuracy differs depending each task. For arithmetic, commonsense, and algorithmic reasoning tasks, accuracy is measured on the correctness of the final answer. Logical reasoning tasks, namely, ProntoQA, accuracy measures the  ability of an LLM to generate the correct reasoning chain. Finally, for planning tasks, accuracy measures the correctness of the proposed plan.

\subsection{Results}
In this subsection, we present the results of the Sys2Bench benchmark, organized by the types of reasoning outlined in Section~\ref{sec:eval}. This grouping allows for a clearer comparison of performance across tasks, demonstrating the strengths and limitations of different inference-time techniques.

\textbf{Arithmetic Reasoning} tasks in Table~\ref{tab:results-big} have strong results with CoT. Performance further improves with SC, as it reduces the impact of randomness in the CoT answers. However, this strong performance does not transfer to tree search methods. ToT significantly underperforms on this task, as its approach of prompting the LLM to explore multiple reasoning paths relies on the LLM generating and selecting correct intermediate reasoning steps. Since LLMs struggle with self-verification~\citep{huang2024large}, it selects incorrect intermediate arithmetic steps, leading to wrong answers. In contrast, RAP shows modest gains on the GSM8K dataset, benefiting from the LLM’s role as a world model to select better arithmetic steps. However, RAP still underperforms SC on AQuA, indicating that tree search methods are not well-suited for arithmetic reasoning tasks. Meanwhile, LRMs deliver exceptional arithmetic reasoning performance, as shown in Table~\ref{tab:o1-results}, highlighting their strength in arithmetic.

\textbf{Logical Reasoning} results in Table~\ref{tab:results-big} show interesting trends. For instance, SC improves performance over CoT on LLaMa 3.1 8B and 70B. However, for LLaMa 3.1 405B and GPT-based models, SC results in performance drops, as it increases the likelihood of generating multiple incorrect reasoning chains in the ProntoQA task, where evaluation focuses on the accuracy of these chains. Majority voting does not help when the LLM outputs multiple wrong reasoning chains. Consistent with arithmetic reasoning, tree search methods such as ToT and RAP also underperform in this task, indicating their limitations in logical reasoning. Finally, as shown in Table~\ref{tab:o1-results}, LRMs do not consistently outperform LLMs on this task, with O1 performing worse than GPT-4o on this task.

\textbf{Common Sense Reasoning} performance of CoT and SC improves with increasing LLM size. However, tree search methods show unique trends. Specifically, both RAP and ToT generate supporting facts for each question, but their effectiveness varies by task. 
To be specific, in StrategyQA, the binary output (yes or no) enables LLaMA models to effectively utilize the generated facts, leading to improved performance. In contrast, for HotPotQA, tree search is not effective as the LLM needs to output short answers. Additional facts often cause LLM hallucinations and increased error rates. Furthermore, compared to other tasks, performance improvements seen with LRMs are limited (see Table~\ref{tab:o1-results}).

\textbf{Algorithmic Reasoning} tasks include the Game of 24 and Binpacking datasets, as described in Section~\ref{sec:eval}. Table~\ref{tab:results-big} shows that both CoT and SC underperform on these tasks. Due to the combinatorial optimization nature of these tasks, that require extensive search, tree search methods perform well on all models, except LLaMa 3.1 8B. The smaller size of LLaMa 3.1 8B limits the model to accurately evaluate and determine the next steps toward a solution. When comparing LLMs to LRMs, results in Table~\ref{tab:o1-results} highlight the potential of O1-mini and O1 in solving NP-Hard and NP-Complete problems, with O1 slightly underperforming O1-mini on Game of 24.

\textbf{Planning} tasks are the most challenging in Sys2Bench. Generally, CoT and SC performance improves with larger model sizes, and SC consistently outperforming CoT.

Tree search methods show mixed results across tasks and models. On smaller models, such as LLaMa 3.1 8B and GPT-4o-mini, ToT shows improvements on tasks like Blocksworld and TripPlan. However for larger models and other tasks, ToT often decreases performance. This is because planning tasks require LLMs to generate actions to solve problems, and incorrect actions can lead to incorrect solutions. Although ToT is intended to help LLMs explore multiple reasoning paths, which in planning means considering different actions, LLMs often fail to generate accurate actions, ultimately reducing performance. The other tree search method, RAP, performs exceptionally well on Blocksworld by leveraging the LLM as a world model to predict future states and rewards.  

Compared to LLMs, LRMs perform significantly better on planning tasks, with O1 achieving near-perfect results on Blocksworld. 
However, the Rubik’s Cube task remains challenging for all methods and models, as it requires advanced spatial reasoning and precise prediction of the consequences of each action. Both LLMs and LRMs currently lack the reasoning capabilities needed for this task, making it out-of-distribution (OOD) for current language models.
\begin{table*}[t]
\centering
\small
\setlength{\tabcolsep}{2pt}
\caption{\small Results of Inference Time Techniques across diverse tasks show that as 
model size increases, performance of CoT (CoT)~\citep{wei2022chain} and Self Consistency (SC)~\citep{wang2023selfconsistency} improves. However, this trend doesn't extend to tree search methods like Tree of Thought (ToT)~\citep{yao2024tree}, where performance does not improve with the bigger models. Furthermore, a comparison between ToT and Reasoning as Planning with World Models (RAP)~\citep{rap} shows that RAP outperforms ToT in planning and arithmetic reasoning tasks but lags in commonsense reasoning while performing equally in algorithmic reasoning tasks. All methods and LLMs fail to solve the Rubik's Cube planning task. This failure can be attributed to the spatial understanding capabilities required for the task, which are currently out of distribution (OOD) for existing LLMs.}
\label{tab:results-big}
\scalebox{0.8}{

\begin{tabular}{@{}ccccccccccccccccccccc@{}}
\specialrule{1.5pt}{0pt}{3pt}
\multicolumn{1}{c}{\multirow{4}{*}{Methods}} & \multicolumn{10}{c}{Algorithmic Reasoning} & \multicolumn{10}{c}{Logical Reasoning} \\ \cmidrule(lr){2-11} \cmidrule(lr){12-21} 
\multicolumn{1}{c}{} & \multicolumn{5}{c}{GSM8K} & \multicolumn{5}{c}{AQuA} & \multicolumn{10}{c}{ProntoQA} \\ \cmidrule(lr){2-6}\cmidrule(lr){7-11}\cmidrule(lr){12-21}  
\multicolumn{1}{c}{} & \multicolumn{3}{c}{LLaMa 3.1} & \multicolumn{2}{c}{GPT} & \multicolumn{3}{c}{LLaMa 3.1} & \multicolumn{2}{c}{GPT} & \multicolumn{6}{c}{LLaMa 3.1} & \multicolumn{4}{c}{GPT} \\ \cmidrule(lr){2-4}\cmidrule(lr){5-6}\cmidrule(lr){7-9} \cmidrule(lr){10-11}\cmidrule(lr){12-17}\cmidrule(lr){18-21}
\multicolumn{1}{c}{} & \multicolumn{1}{c}{8B} & \multicolumn{1}{c}{70B} & \multicolumn{1}{c}{405B} & \multicolumn{1}{c}{4o mini} & \multicolumn{1}{c}{4o} & \multicolumn{1}{c}{8B} & \multicolumn{1}{c}{70B} & \multicolumn{1}{c}{405B} & \multicolumn{1}{c}{4o mini} & \multicolumn{1}{c}{4o} & \multicolumn{2}{c}{8B} & \multicolumn{2}{c}{70B} & \multicolumn{2}{c}{405B} & \multicolumn{2}{c}{4o mini} & \multicolumn{2}{c}{4o} \\ 
\midrule
\multicolumn{21}{c}{Chain of Thought Methods} \\ \midrule
\multicolumn{1}{l|}{CoT} & 79.8 & 95.5 & 97.0 & 92.6 & 94.7 & 58.7 & 77.2  & 78.0  & 73.6 & 79.9 & \multicolumn{2}{c}{45.8} & \multicolumn{2}{c}{82.6} & \multicolumn{2}{c}{91.0} & \multicolumn{2}{c}{61.4} & \multicolumn{2}{c}{91.8} \\
\multicolumn{1}{l|}{SC @ 5} & 86.7 & 96.5 & 97.5 & 93.3 & 94.9 & 70.9 & 85.8 & 86.2 & 79.9 & 83.9 & \multicolumn{2}{c}{54.2} & \multicolumn{2}{c}{88.4} & \multicolumn{2}{c}{89.0} & \multicolumn{2}{c}{58.0} & \multicolumn{2}{c}{91.4} \\ 
\midrule\multicolumn{21}{c}{Tree Search Methods} \\ \midrule
\multicolumn{1}{l|}{ToT} & 60.0 & 91.5 & 96.0 & 91.5  & 93.5 & 44.8 & 78.0 & 85.8 & 81.1 & 78.0 & \multicolumn{2}{c}{13.5} & \multicolumn{2}{c}{24.2} & \multicolumn{2}{c}{62.6} & \multicolumn{2}{c}{42.0} & \multicolumn{2}{c}{32.8} \\
\multicolumn{1}{l|}{RAP} & 87.3 & -- & -- & -- & -- & 68.1 & -- & -- & -- & -- & \multicolumn{2}{c}{0.0} & \multicolumn{2}{c}{--} & \multicolumn{2}{c}{--} & \multicolumn{2}{c}{--} & \multicolumn{2}{c}{--} \\ \specialrule{1.2pt}{0pt}{3pt}
 & \multicolumn{10}{c}{Common Sense Reasoning} & \multicolumn{10}{c}{Algorithmic Reasoning} \\ \cmidrule(lr){2-11} \cmidrule(lr){12-21} 
 & \multicolumn{5}{c}{HotPotQA} & \multicolumn{5}{c}{StrategyQA} & \multicolumn{5}{c}{Game of 24} & \multicolumn{5}{c}{Binpacking} \\ \cmidrule(lr){2-6}\cmidrule(lr){7-11} \cmidrule(lr){12-16}\cmidrule(lr){17-21}
 & \multicolumn{3}{c}{LLaMa 3.1} & \multicolumn{2}{c}{GPT} & \multicolumn{3}{c}{LLaMa 3.1} & \multicolumn{2}{c}{GPT} & \multicolumn{3}{c}{LLaMa 3.1} & \multicolumn{2}{c}{GPT} & \multicolumn{3}{c}{LLaMA 3.1} & \multicolumn{2}{c}{GPT} \\ \cmidrule(lr){2-4}\cmidrule(lr){5-6}\cmidrule(lr){7-9}\cmidrule(lr){10-11}\cmidrule(lr){12-14}\cmidrule(lr){15-16}\cmidrule(lr){17-19}\cmidrule(lr){20-21} 
 & \multicolumn{1}{c}{8B} & \multicolumn{1}{c}{70B} & \multicolumn{1}{c}{405B} & \multicolumn{1}{c}{4o mini} & \multicolumn{1}{c}{4o} & \multicolumn{1}{c}{8B} & \multicolumn{1}{c}{70B} & \multicolumn{1}{c}{405B} & \multicolumn{1}{c}{4o mini} & \multicolumn{1}{c}{4o} & \multicolumn{1}{c}{8B} & \multicolumn{1}{c}{70B} & \multicolumn{1}{c}{405B} & \multicolumn{1}{c}{4o mini} & \multicolumn{1}{c}{4o} & \multicolumn{1}{c}{8B} & \multicolumn{1}{c}{70B} & \multicolumn{1}{c}{405B} & \multicolumn{1}{c}{4o mini} & \multicolumn{1}{c}{4o} \\ 
 \midrule
 \multicolumn{21}{c}{Chain of Thought Methods} \\ \midrule
\multicolumn{1}{l|}{CoT} & 13.8 & 30.6 & 41.0 & 38.6  & 52.8 & 46.0 & 61.5 & 76.0 & 76.6 & 79.2 & 6.0 & 8.0 & 7.0 & 13.0 & 14.0 & 6.0 & 33.0 & 45.0 & 31.0 & 75.0 \\
\multicolumn{1}{l|}{SC @ 5} & 20.6 & 36.6 & 45.6 & 40.6 & 52.6 & 53.5 & 66.0 & 78.5 & 76.0 & 79.8 & 6.0 & 8.0 & 6.0 & 15.0 & 18.0 & 6.0 & 45.0 & 64.0 & 41.0 & 86.0 \\
\midrule\multicolumn{21}{c}{Tree Search Methods} \\ \midrule
\multicolumn{1}{l|}{ToT} & 23.0 & 30.0 & 31.5 & 31.4 & 38.2 & 68.0 & 82.0 & 79.5 & 67.5 & 73.5 & 1.0 & 59.0 & 69.0 & 42.0 & 62.0 & 1.0 & 46.0 & 81.0 & 53.0 & 77.0 \\
\multicolumn{1}{l|}{RAP} & -- & -- & -- & -- & -- & 58.5 & -- & -- & -- & -- & 1.0 & -- & -- & -- & -- & 1.0 & -- & -- & -- & -- \\ \specialrule{1.2pt}{0pt}{3pt}
 & \multicolumn{20}{c}{Planning} \\ \cmidrule(lr){2-21} 
 & \multicolumn{5}{c}{Blocksworld} & \multicolumn{5}{c}{Trip Plan} & \multicolumn{5}{c}{Calendar Plan} & \multicolumn{5}{c}{Rubik's Cube} \\ \cmidrule(lr){2-6}\cmidrule(lr){7-11} \cmidrule(lr){12-16}\cmidrule(lr){17-21}
 & \multicolumn{3}{c}{LLaMa 3.1} & \multicolumn{2}{c}{GPT} & \multicolumn{3}{c}{LLaMa 3.1} & \multicolumn{2}{c}{GPT} & \multicolumn{3}{c}{LLaMa 3.1} & \multicolumn{2}{c}{GPT} & \multicolumn{3}{c}{LLaMa 3.1} & \multicolumn{2}{c}{GPT} \\ \cmidrule(lr){2-4}\cmidrule(lr){5-6}\cmidrule(lr){7-9}\cmidrule(lr){10-11}\cmidrule(lr){12-14}\cmidrule(lr){15-16}\cmidrule(lr){17-19}\cmidrule(lr){20-21}  
 & \multicolumn{1}{c}{8B} & \multicolumn{1}{c}{70B} & \multicolumn{1}{c}{405B} & \multicolumn{1}{c}{4o mini} & \multicolumn{1}{c}{4o} & \multicolumn{1}{c}{8B} & \multicolumn{1}{c}{70B} & \multicolumn{1}{c}{405B} & \multicolumn{1}{c}{4o mini} & \multicolumn{1}{c}{4o} & \multicolumn{1}{c}{8B} & \multicolumn{1}{c}{70B} & \multicolumn{1}{c}{405B} & \multicolumn{1}{c}{4o mini} & \multicolumn{1}{c}{4o} & \multicolumn{1}{c}{8B} & \multicolumn{1}{c}{70B} & \multicolumn{1}{c}{405B} & \multicolumn{1}{c}{4o mini} & \multicolumn{1}{c}{4o} \\ \midrule
 \multicolumn{21}{c}{Chain of Thought Methods} \\ \midrule
\multicolumn{1}{l|}{CoT} & 3.5 & 26.1 & 48.7 & 18.4  & 37.5 & 12.3 & 29.5 & 27.0 & 5.3 & 6.3 & 10.4 & 31.2 & 44.8 & 26.0 & 47.0 & 0.6 & 0.0 & 0.0 & 0.6 & 0.0 \\
\multicolumn{1}{l|}{SC @ 5} & 4.5 & 30.7 & 52.1 & 21.2 & 41.5 & 12.0 & 32.3 & 34.3 & 5.0 & 5.8 & 11.6 & 38.0 & 45.6 & 29.6 & 47.4 & 0.0 & 0.6 & 0.6 & 0.6 & 0.6 \\
\midrule\multicolumn{21}{c}{Tree Search Methods} \\ \midrule
\multicolumn{1}{l|}{ToT} & 13.9 & 4.6 & 19.9 & 23.1 & 12.4 & 2.0 & 32.5 & 29.5 & 7.8 & 19.5 & 16.8 & 32.0 & 40.0 & 29.0 & 41.4 & 0.6 & 0.6 & 0.6 & 0.0 & 0.6 \\
\multicolumn{1}{l|}{RAP} & 46.8 & -- & -- & -- & -- & -- & -- & -- & -- & -- & -- & -- & -- & -- & -- & 0.6 & -- & -- & -- & -- \\
\specialrule{1.5pt}{0pt}{3pt}
\end{tabular}

}
\end{table*}

\begin{table*}[t]
\small
\centering
\caption{\small Results of large reasoning models (LRMs). We report the results of IO prompting on LRMs, including OpenAI O1-mini and O1, without providing any in-context learning examples, as recommended by OpenAI~\citep{openai2024o1systemcard}. Overall, LRMs achieve state-of-the-art performance, with O1 outperforming O1-mini on all tasks except the Game of 24. Similar to LLMs, LRMs also struggle with the Rubik's Cube task, indicating a lack of spatial understanding. }
\label{tab:o1-results}

\setlength{\tabcolsep}{2pt}  
\scalebox{0.75}{ 
\begin{tabular}{@{}ccccccccccccc@{}}
\toprule
\multirow{2}{*}{} &
  \multicolumn{2}{c}{\begin{tabular}[c]{@{}c@{}}Arithmetic \\ Reasoning\end{tabular}} &
  \multicolumn{1}{c}{\begin{tabular}[c]{@{}c@{}}Logical \\ Reasoning\end{tabular}} &
  \multicolumn{2}{c}{\begin{tabular}[c]{@{}c@{}}Common Sense \\ Reasoning\end{tabular}} &
  \multicolumn{2}{c}{\begin{tabular}[c]{@{}c@{}}Algorithmic \\ Reasoning\end{tabular}} &
  \multicolumn{4}{c}{Planning} \\ 
  \cmidrule(lr){2-3} \cmidrule(lr){4-4} \cmidrule(lr){5-6} \cmidrule(lr){7-8} \cmidrule(lr){9-12} 
        & GSM8K & AQuA & ProntoQA & HotPotQA & StrategyQA & Game of 24 & Binpacking & Blocksworld & Trip Plan & Calendar Plan & Rubik's Cube \\ 
        \cmidrule(lr){1-12} 
O1 Mini & 98.0 & 92.0 & 64.0 & 35.0 & 74.0 &  77.0 & 90.0 & 48.3 & 24.0 & 88.2 &  0.0  \\
O1      & 98.0 & 91.0 & 74.0 & 59.0 & 81.0 & 73.0  & 99.0 & 99.2 & 58.3 & 90.0 & 0.6  \\ 
\bottomrule
\end{tabular}
}
\end{table*}

\subsection{Insights}
We extend our main experiments to provide additional insights and uncover important trends. As the research community shows increasing interest in inference-time techniques and improving LLM reasoning, these findings offer valuable contributions to ongoing discussions.

\textbf{Inference-time compute scaling is limited by LLM bias}. These techniques aim to improve LLM reasoning by guiding them to generate intermediate steps, simplifying complex tasks into smaller, manageable parts. However, this premise is flawed as LLMs do not exhaustively search for all reasoning paths and remain biased toward certain ones. As inference-time compute scales, this bias persists, limiting exploration and leading to diminished performance. As task complexity increases, this issue becomes worse, exacerbating errors in reasoning and decision-making.

Our Sys2Bench experiments show this trend in arithmetic and logical reasoning. In these tasks, LLMs excel with CoT but struggle with tree search, failing to explore reasoning paths and select the correct one. 


\begin{figure}[t]
    \centering
    \begin{minipage}{0.49\linewidth}
        \centering
        {\footnotesize (a) Trip Plan} \\ 
        \vspace{2mm} 
        \includegraphics[width=\linewidth, clip=true,trim = 0mm 0mm 0mm 0mm]{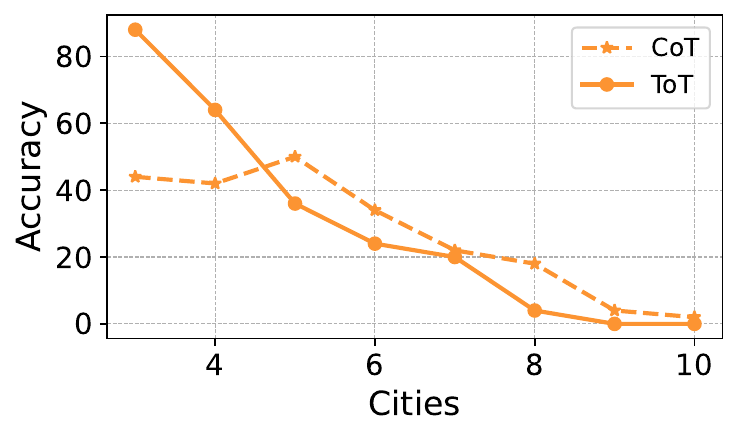}
    \end{minipage}
    \hfill
    \begin{minipage}{0.49\linewidth}
        \centering
        {\footnotesize(b) Blocksworld} \\ 
        \vspace{2mm} 
        \includegraphics[width=\linewidth, clip=true,trim = 0mm 0mm 0mm 0mm]{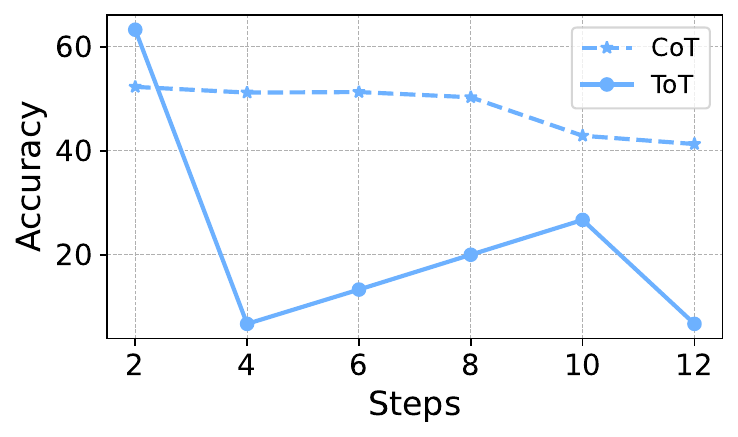}
    \end{minipage}
    
    \caption{\small Tree of Thought (ToT) performance declines as task complexity increases. In (a) Trip Plan and (b) Blocksworld, the number of steps or cities represents the required tree depth for LLM inference-time search. While ToT performs well for smaller depths, performance deteriorates as problem complexity grows, eventually falling below CoT. Notably, CoT achieves better performance with significantly lower computational resources, as shown in Table~\ref{tab:compute}. These results are on the LLaMa 405B model.} 
    \label{fig:tree_depth}
\end{figure}

\textbf{Tree search struggles with increasing complexity, performing significantly worse than CoT.} As shown in Fig~\ref{fig:tree_depth}, its benefits diminish beyond a depth of 4 for the TripPlanning and Blocksworld tasks on LLaMa 3.1 405B. Note that, LLaMa 3.1 405B has a strong CoT performance in challenging planning tasks and ideally ToT should lead to further improvements. However, as complexity grows, generating the right intermediate steps becomes crucial, leading to worse performance of ToT compared to CoT. A potential explanation for this observation is the inherent bias of LLMs at each step of the reasoning process. These biases may propagate through successive steps, leading to cumulative errors that degrade ToT performance.\\

\begin{figure}[t]
    \centering
    \begin{minipage}{0.48\linewidth} 
        \centering
        \vspace{0pt}
        \includegraphics[width=\linewidth, clip=true, trim=0mm 0mm 0mm 0mm]{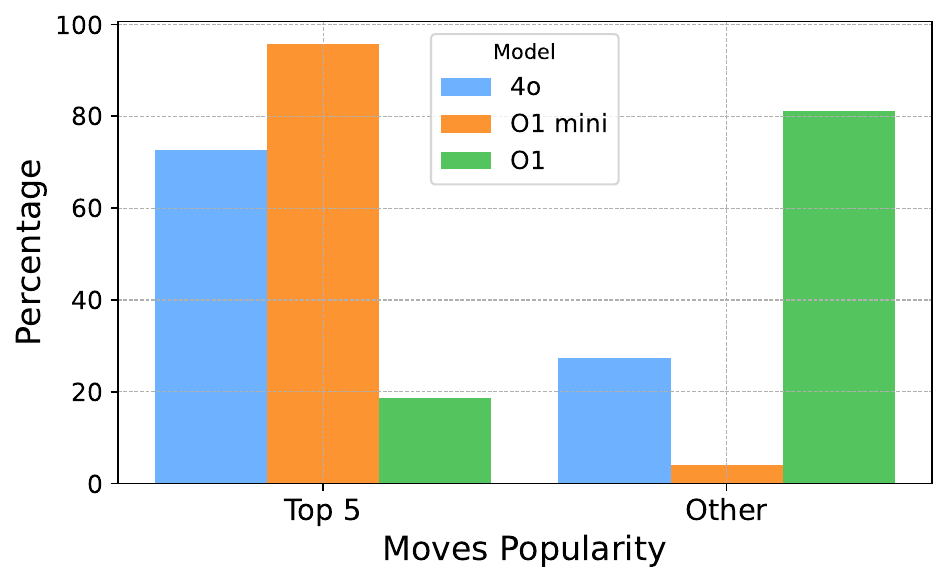}
        \caption{\small Moves are grouped based on their popularity in online Rubik's Cube solutions. LLMs like GPT-4o frequently generate moves commonly found in online algorithms. This issue is even more pronounced in O1-mini, where nearly 90\% of the moves are repeated! O1 exhibits this behavior less frequently, but it remains a notable pattern.} 
        \label{fig:cube}
    \end{minipage}%
    \hfill
    \begin{minipage}{0.48\linewidth} 
        \centering
        \vspace{-2cm}
        \captionof{table}{\small Token count comparison of CoT, SC, ToT, and RAP on LLaMA 3.1 8B across Blocksworld, Game of 24, and GSM8K. The results highlight that scaling up inference-time techniques increases computational cost without proportionate performance gains, contrasting with the trends observed by~\citet{snell2024scaling}.}
        \label{tab:compute}
        \scalebox{0.66}{
        \begin{tabular}{@{}ccccccccc@{}}
        \toprule
        \multirow{2}{*}{} &
          \multicolumn{2}{c}{Blocksworld} &
          \multicolumn{2}{c}{Game of 24} &
          \multicolumn{2}{c}{GSM8K} \\ \cmidrule(lr){2-3} \cmidrule(lr){4-5} \cmidrule(lr){6-7}
                & Acc $\uparrow$ & Tokens $\downarrow$ & Acc $\uparrow$& Tokens $\downarrow$ & Acc $\uparrow$ & Tokens $\downarrow$ \\ \midrule
        CoT     & 3.5      & $3.2\times10^4$   & 6.0        & $8.0\times10^3$     & 79.8     & $1.9\times10^4$     \\ 
        SC      & 4.5      & $1.7\times10^5$  & 6.0        & $4.0\times10^4$    & 86.7     & $2.4\times10^5$    \\ 
        ToT     & 13.9     & $1.1\times10^6$ & 1.0        & $3.6\times10^7$ & 60.0     &  $4.4\times10^6$   \\ 
        RAP     & 46.8     & $4.9\times10^5$  & 1.0        & $1.5\times10^7$ & 87.3     & $9.9\times10^6$   \\ \bottomrule
        \end{tabular}}
    \end{minipage}
\end{figure}

\textbf{Language models rely on retrieval rather than true understanding}. 
Despite advancements in reasoning abilities with LRMs such as O1 and O1-Mini, they still appear to be pattern matching rather than genuine reasoning. This issue has been observed in prior studies for LLMs \citep{valmeekam2023planning}, but we are the first to demonstrate it for LRMs, including O1 and O1-Mini. 

As shown in Fig.~\ref{fig:cube}, we compare GPT-4o, O1-Mini, and O1 based on how frequently their generated moves in the Rubik's Cube task align with known online algorithms. Our analysis shows that GPT-4o and O1-Mini repeat these popular moves in nearly all cases, with rates of 75\% and 90\%, respectively. While O1 performs better, it still follows common move sequences about 20\% of the time.

\textbf{There is a tradeoff between performance and the cost of inference-time methods}. Table~\ref{tab:compute} includes results across Blocksworld, Game of 24, and GSM8K using LLaMa 3.1 8B, along with the number of tokens generated per task. 
Compared to CoT and SC, ToT and RAP have significantly higher computational costs. However, increased token usage does not mean better performance. Additionally, solving 100 Game of 24 problems with GPT-4o and ToT costs around \$60 due to high token usage, with costs rising for larger models and harder tasks. This tradeoff underscores that increasing inference-time computation does not necessarily translate to proportional improvements in performance. 

\section{Discussions}
While our study highlights certain limitations, there is growing interest in scaling inference-time computation to further enhance LLM performance~\citep{snell2024scaling}. These methods have demonstrated strong results in arithmetic~\citep{kumar2024traininglanguagemodelsselfcorrect} and gameplay tasks~\citep{schultz2024masteringboardgamesexternal}. However, these improvements rely on verifiers or external models to guide reasoning, and without them, performance gains disappear. Moreover, tasks like Common Sense Reasoning lack verifiers, higlighting the limitation of verifier-dependent inference-time scaling.

On the other hand, LRMs like O1 leverage reinforcement learning (RL) with inference-time search to improve reasoning~\citep{openai2024o1systemcard, zhao2024marcoo1openreasoningmodels}. These models are trained to generate correct reasoning steps, enabling better performance (see Table~\ref{tab:results-big}). Despite significant gains, our experiments highlight limitations, particularly in planning and common sense reasoning tasks. With inference costs reaching $60\times$ that of standard LLMs, it is crucial to assess their limitations.

In contrast to alternative views, we argue that simply scaling inference-time computation is not the solution. Instead, improving LLM reasoning requires a more strategic approach, and combination of RL with inference-time methods has been promising~\citep{deepseek}. The recent release of DeepSeek-R1 hints at progress in this direction, offering a glimpse of what more refined models could achieve in the future.

\section{Conclusion}
This paper examines the impact of scaling inference-time computation on improving the reasoning and planning abilities of LLMs. We show that scaling inference-time computation has limitations. Instead, we need
to explore diverse approaches to enhance the holistic
reasoning capabilities of LLMs. We explore this by introducing Sys2Bench, a new benchmark, and conduct extensive experiments evaluating inference-time techniques across eleven diverse tasks spanning five categories, namely, arithmetic reasoning, logical reasoning, common sense reasoning, algorithmic reasoning, and planning. Our findings provide important 
insights into the limitations of inference-time techniques. Finally, we discuss alternative perspectives from the literature, critically analyze their implications, and outline potential directions for future research.

\section*{Acknowledgments}

This work was supported in part by National Institutes of Health under grant
U01AG070112 and National Science Foundation under grant
CNS-2328395.

{
\small
\bibliographystyle{plainnat}
\bibliography{sys2bench_neurips_data_2024}
}

\end{document}